\pdfoutput=1
\documentclass[manuscript, authorversion, nonacm]{acmart}

\AtBeginDocument{%
  \providecommand\BibTeX{{%
    \normalfont B\kern-0.5em{\scshape i\kern-0.25em b}\kern-0.8em\TeX}}}

\setcopyright{acmcopyright}
\copyrightyear{2023}
\acmYear{2023}
\acmDOI{XXXXXXX.XXXXXXX}

\DeclareMathOperator*{\argmin}{arg\,min} 
\usepackage{booktabs}
\begin{document}

\title[AI Art Curation]{AI Art Curation: Re-imagining the city of Helsinki in occasion of its Biennial}
\author{Ludovica Schaerf}
\authornotemark[1]
\orcid{0000-0001-9460-702X}
\affiliation{%
  \institution{Max Planck Society – University of Zurich}
  \streetaddress{Center for Digital Visual Studies -Culmannstrasse 1}
  \city{Zurich}
  \state{}
  \country{Switzerland}
  \postcode{8006}
}
\email{ludovica.schaerf@uzh.ch}

\author{Pepe Ballesteros Zapata}
\authornotemark[1]
\affiliation{%
  \institution{Max Planck Society – University of Zurich}
 \country{Switzerland}
}
\email{jose.ballesteroszapata@uzh.ch}

\author{Valentine Bernasconi}
\authornotemark[1]
\affiliation{%
  \institution{Max Planck Society – University of Zurich}
  \country{Switzerland}
}
\email{valentine.bernasconi@uzh.ch}

\author{Iacopo Neri}
\authornotemark[1]
\affiliation{%
 \institution{Max Planck Society – University of Zurich}
 \country{Switzerland}
  }
\email{neri.iacopo@gmail.com}

\author{Dario Negueruela del Castillo}
\authornote{All authors contributed equally to the paper}
\affiliation{%
  \institution{Max Planck Society – University of Zurich}
  \country{Switzerland}
}
\email{dario.neguerueladelcastillo@uzh.ch}

\renewcommand{\shortauthors}{Schaerf et al.}

\begin{abstract}
 \textbf{Abstract.}  Art curatorial practice involves presenting an art collection in a knowledgeable way. Machine processes, on the other hand, are characterized by their ability to manage and analyze vast amounts of data. This paper explores the implications of contemporary machine learning models for the curatorial world through AI curation and audience interaction. The project was developed for the 2023 Helsinki Art Biennial, titled "New Directions May Emerge," and utilizes the Helsinki Art Museum (HAM) collection to re-imagine the city of Helsinki through the lens of machine perception. Visual-textual models are used to place museum artworks in public spaces, assigning fictional coordinates based on similarity scores. They are then used to generate synthetic 360° art panoramas, transforming the space that each artwork inhabits in the city. The generation is guided by estimated depth values from 360° panoramas at each artwork location and using machine-generated prompts of the artworks. The result is an AI curation that places the artworks in their imagined physical space, blurring the lines of artwork, context, and machine perception. The project is virtually presented as a web-based installation, where users can navigate an alternative version of the city and explore and interact with its cultural heritage at scale.
\end{abstract}



\begin{teaserfigure}
  \centering
  \includegraphics[width=\textwidth]{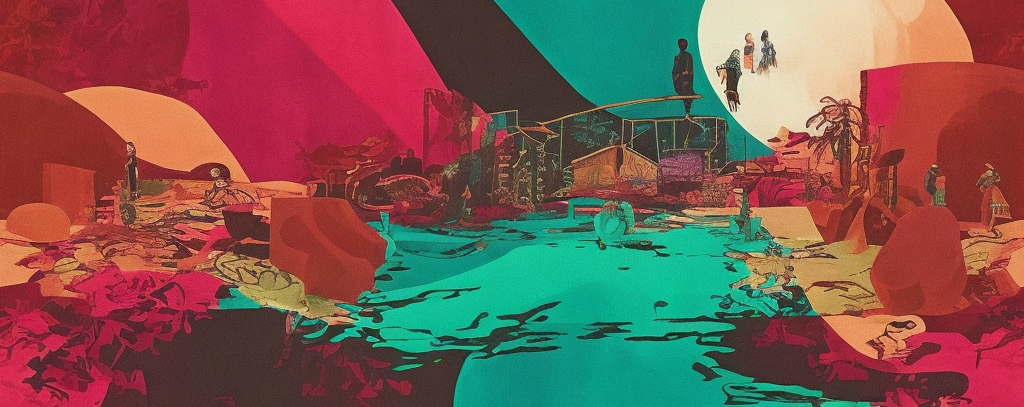}
  \caption{Example of a 360° art panorama. A real location from Helsinki is transformed according to the artistic style of its corresponding artwork.}
  \Description{Panoramic image of a HAM artwork in the city.}
  \label{fig:teaser}
\end{teaserfigure}


\maketitle

\section{Introduction}

In this paper, we present a curatorial work that was exhibited on the occasion of the 2023 Helsinki Biennial of Art. The project uses Artificial Intelligence (AI) as a new means for curatorial practice, exploring the possibilities and difficulties that such new methods introduce. The proposed work in this paper is part of a more significant project entitled \textit{New Directions May Emerge}\footnote{See \url{https://helsinkibiennaali.fi/en/story/helsinki-biennial-2023-brings-together-29-artists-and-collectives/}.}, which aims to curate a museum’s art collection through the perception of the machine. The immediate product of this curation consists of an interactive website\footnote{\url{http://newlyformedcity.net/}} that transforms the shape of Helsinki using the virtual dots of the collection locations. The website dynamically updates throughout the Biennial, with new locations (and therefore artworks) appearing and modifying the shape of the city every 30 minutes. 

The work is based on the art collection from the Helsinki Art Museum (HAM), consisting of public artworks, such as sculptures and art installations, as well as an indoor collection. Using the city of Helsinki as a context, the goal is to present a different experience of the works of art through the navigation of a new projection of the artworks in the city. First, we situate the artworks from the indoor museum collection to a fictional place in Helsinki, using deep learning and machine learning tools. We then extract the 360° Google Maps views of all the real and fictional locations, which are later used as guiding depth maps to embed the artworks in their fictional surrounding space. This is achieved using Stable Diffusion \cite{Rombach2022-va} on the 360° views that they inhabit. The city is thereafter populated by the new machinic world, where the user can navigate a geography that blurs the world of extant reality and that of machinic fiction. 
Following the narrative thread proposed in \cite{Krysa2014-pf}, the project focuses on the following questions: How can we curate a collection we have never seen? How does the machine perceive art? Can the machine offer a fruitful re-contextualization of the artistic data? Can the geography of the city offer fruitful ground for this re-contextualization?

The contributions of this paper can be summarized as follows: 1) We provide a survey of existing AI curatorial methods and situate this novel avenue in the context of digital and physical curation, 2) We present an innovative method for AI curation that explores the connection to the physical setting and offers a re-contextualization of the original art, 3) We introduce a comprehensive evaluation procedure which takes into consideration both the technical quality of the final artworks and the artistic licenses, 4) We thoroughly discuss the social and ethical implications of our project and, generally, AI curation.  

\subsection{AI as a Curator}
Traditionally, the work of a museum curator consists of the enrichment of a collection and cultural heritage preservation. With the creation of an exhibition, the curator performs a selection of the collection according to a narrative thread that has to be passed to the public \cite{Heinich1989-ck}. The goal is to generate new insights into the original works of art, and elevate their physical dimension through the design of their display \cite{Davis1980-io}. In recent years, with the increased availability of digital collections and tools, the notion of digital curation has become an important aspect \cite{Poole2016-gj}, especially facing the large amount of digital data generated and their online publication to reach a wider audience. Computational art curation aims at classifying and indexing data for efficient retrieval \cite{Poole2016-gj}, as well as creating new experiences of the artworks through new technologies \cite{Giannachi_undated-pu} (e.g., virtual and augmented reality).

Unsurprisingly, the use of AI systems for artistic curation has found fertile ground, spanning from projects by Google Arts \& Culture to on-site curations of museums and biennials. Google Arts \& Culture’s\footnote{See \url{https://experiments.withgoogle.com/collection/arts-culture}.} experiments are pioneering applications of computational methods to the online curation of artistic datasets. For example, their 't-SNE Map' and 'Curator Table' experiments are visualization tools to see how objects, styles, and artists evolve over time. Moreover, the project ‘X degrees of separation’ was a source of inspiration for the AI-based curatorial project \textit{Dust and Data: The Art of Curating in the Age of Artificial Intelligence} \cite{Flexer2021-gj}. 

In fact, \textit{Dust and Data} (DAD) explores the possibilities of AI curation as an assistant to both curators and audiences; it uses semantic embeddings to recreate a curatorially specific version of the Google Arts experiment, which proposes a chain of artworks that connect one work to another, in this sense, filling the curatorial gaps in art collections \cite{Flexer2021-gj, Flexer_undated-hz}.

Another remarkable example of machine curation is the previous edition of a Biennial, the Liverpool Biennial 2021, titled \textit{The Next Biennial Should be Curated by a Machine}\footnote{See \url{https://ai.biennial.com/}.}. The curation allows navigating the Liverpool collection through a set of alien images. For each artwork, GAN-generated images were created from the titles of existing artworks in the collections. Moreover, CLIP \cite{Radford2021-va} was used to extract keywords from the artworks that were used as the link to navigate the collection. 

AI curation aims to offer new insights into digital cultural artifacts. It is possible to propose personalized journeys of the collections, as well as to foster creative takes on its presentation \cite{Giannachi_undated-pu}. Finally, contemporary AI curation strives to disentangle the undercovered behaviors of large models by switching from the practical data and tasks they were trained on to curatorial and artistic purposes.

In this field, AI technology and its implementations are evolving steadily fast. The incorporation of physical space has already been explored in interactive place-making with the use of data curation technologies like recommender systems [3]. Our incorporation of references to real locations and the generation of new imagined immersive panoramas make this work novel in regards to previous and current efforts in urban curation using AI, whilst other projects have been limited to the visual and textual dimensions. Specifically, regarding the general trends and approaches identified by [2]. In addition, our approach to first interpret HAM artworks textually through CLIP engages with the discursive practices of art curation and the necessary critique and self-reflective practice in digital curation as articulated by Cruz [4]. It is crucial to distinguish between the different literature on AI and the digital curation of cultural and artistic data. On the one hand, that belonging to the contemporary digital art curating practices, and on the other, that focused on cultural heritage and its management, usually more focused on infrastructural solutions and less on discursive practices, as summarized by [1]. In the literature, a general concern regards the condition of AI as a ‘black-box’ and its effects of disempowerment of a passive audience. Even though our work has an artistic context that is not primarily nor solely concerned with explainable AI, we will also clearly characterize our work with respect to this.

The way we incorporate the urban context is still mainly limited to visual inputs (360° panoramas) but we do start to imply the crucial third dimension through the depth estimation using MiDaS. In addition, in contrast to other similar projects like “The Next Biennial Should Be Curated By a Machine”[6], we also include spatial contextual information in the process of guiding the generation of new artistic panoramas, and we do not simply restrain ourselves to the outputs of an abstract visual and textual latent space. In this respect, special attention needs to be put into understanding the place of this project and methodology with regard to curating cities with the aid of computational methods and machine learning in particular. 

A crucial element relevant to ethical considerations of AI curation is the possibility of offering a gaze on any art collection that would be free from a specific cultural framing. Indeed, as pointed out by Jones \cite{Jones1993-sq}, there has been over the past four decades an increasing criticism of the way cultural objects that do not belong to our Western culture are treated. These objects are too often perceived as primitive artifacts, which derive from a colonial projection on non-western societies \cite{Amselle2003-od, Jones1993-sq}. The approach of the museum curator towards the objects and the narrative presented in an exhibition impact and influence the perception of the public, and solutions have to be found to overcome this biased gaze based on the origin of artworks, and to open to “\textit{alternative voices, histories, and representations}” \cite{Jones1993-sq}. Nevertheless, considering the material used in the training of most deep learning models and its strong Western anchoring, AI cannot be considered in itself as the solution but as another possibility for experiments toward cultural diversity and postcolonial views.

\subsection{The HAM Dataset}

The Helsinki Art Museum (HAM)’s collection consists of the core material used for the project. Defining itself as “\textit{a city-wide art museum}” the HAM holds about 10’000 artworks, of which around 2’500 can be found in the outdoor and indoor public spaces of the city. These artworks are very diverse, such as sculptures, paintings, and drawings. The idea behind the project is to take advantage of that urban perspective and experiment with the original locations of the public works. To this end, we get access to the geographical information and the corresponding photographs of the 488 outdoor public artworks. The information consists of a set of longitude and latitude coordinates. Additionally, 1’744 items from their indoor collection are harvested from their online platform\footnote{See \url{https://ham.finna.fi/?lng=en-gb} for the full collection.}. For each item, a corresponding image representing the artwork is retrieved, as well as the title, date of creation, name of the artist, keywords in English, Finnish, and Swedish describing the piece, and the object ID in the official collection.

We thus collect a total of 2’232 items divided into two distinctive sets referred to as the public art, corresponding to the outdoor public artworks, and the indoor art, referring to the indoor collection of the HAM.

\section{Methods}

In this project, we strive to present and recreate the HAM collection as a new entity inhabiting and embodying the city of Helsinki. To this end, our curatorial pipeline begins with the geolocation of all the artworks of the collection, including the indoor artworks that do not have a physical location. We proceed in two steps: we employ an image-to-text model to extract a compressed representation of both the public and indoor art, and we successively exploit this representation to assign fictional coordinates to the indoor collection based on their similarity to the public artworks. Given the new coordinates, we proceed to induce the artworks to embody their space at that coordinate: We extract the panoramic 360° view of each artwork from its corresponding location and use diffusion-based models \cite{Rombach2022-va} to turn the 360° panoramas into an immersive space representing the artwork. The output image is generated using depth images of extracted panoramas and machine-generated prompts as input guidance for the model (Figure \ref{fig:6}).

\subsection{Image to CLIP representations}
\begin{figure}[h!]
  \includegraphics[width=0.5\textwidth]{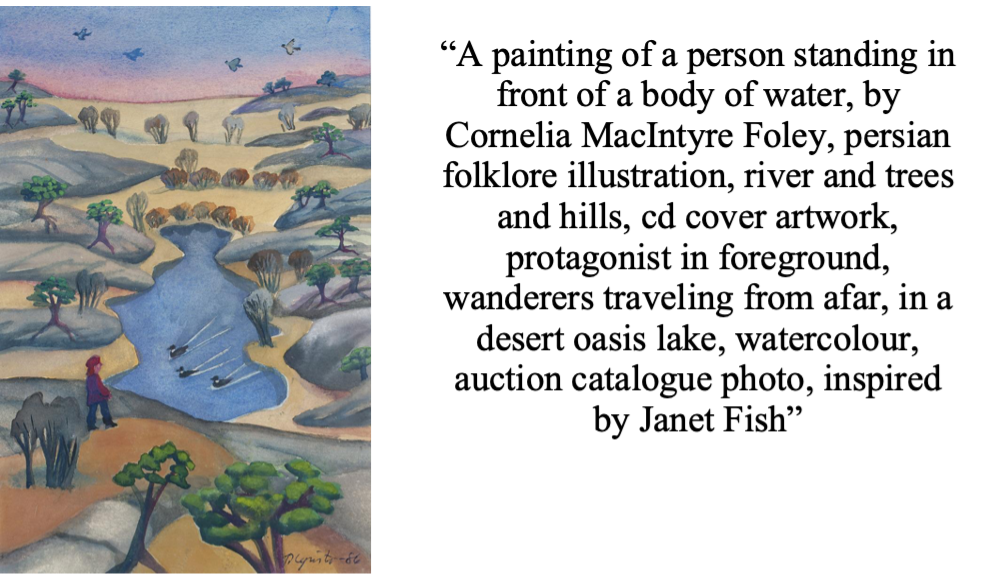}
  \caption{Example artwork with CLIP Interrogator extracted text. Artwork: Lepistö, P. "Kolme vesilintua metsälammella". Courtesy of the HAM collection.
}
  \Description{}
  \label{fig:1}
\end{figure}

As a first step, we extract the visual and textual features from all images in the collection using the CLIP-based model CLIP-Interrogator \cite{Radford2021-va} (Figure \ref{fig:1}). In fact, we take advantage of the zero-shot performance of the CLIP model released by OpenAI, which produces Stable Diffusion 1.5 compatible prompts. Using CLIP-Interrogator, we store two outputs: the prompts $\mathbf{t}$ and the embeddings $\mathbf{z}^*$. For each image $\mathbf{x}\in \mathcal{X}$, the interrogator maps $x$ to the image embedding $\mathbf{z}_I \in \mathcal{Z}_I$ using the \texttt{ViT-L-14} model. It leverages contrastive-based learned features to map image embeddings $\mathbf{z}_I$ to text embeddings $\mathbf{z}_T \in \mathcal{Z}_T$, where $\mathbf{z}_I, \mathbf{z}_T \in \mathbb{R}^m, m = 768$. Each text embedding $\mathbf{z}_T$ is decoded into a text prompt $\mathbf{t} \in \mathcal{T}$. The prompts will be used in Section 2.4 as the inputs for the Stable Diffusion generation. We wish to consider both linguistic and visual information to assign the fictional coordinates in Section 2.2. Therefore, we represent each artwork (both indoor and public) as the concatenation $\mathbf{z}^* \in \mathcal{Z}^*$ of $\mathbf{z}_I$ and $\mathbf{z}_T$, where $\mathbf{z}^* \in \mathbb{R}^{m+m}$.

\subsection{CLIP to Fictional Coordinates}
\begin{figure}[h!]
  \includegraphics[width=0.5\textwidth]{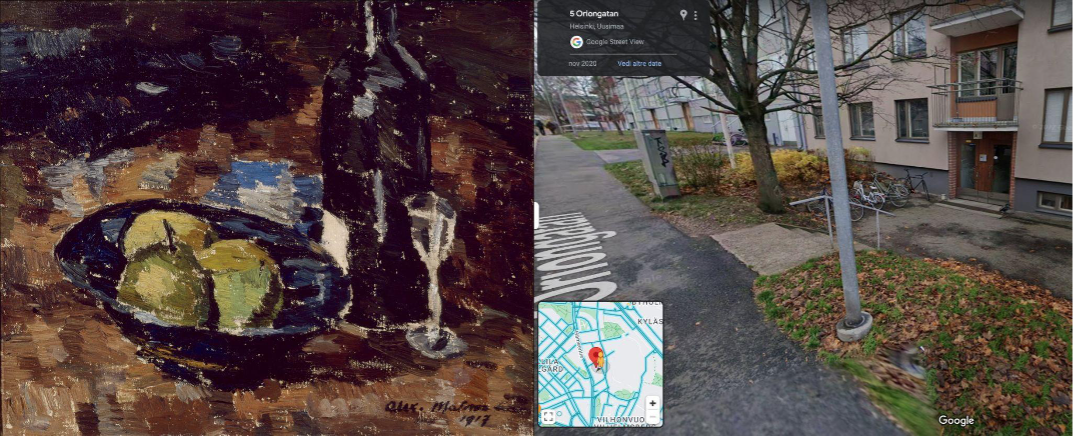}
  \caption{Example indoor artwork (left) with predicted location (right). Artwork: Matson, A. "Asetelma". Courtesy of the HAM collection.
}
  \Description{}
  \label{fig:2}
\end{figure}

Successively, we determine fictional coordinates (Figure \ref{fig:2}) for the 1’744 images of the indoor collection using information from the known geolocations (latitudes and longitudes) of the public artworks $\mathbf{y}_{\mathbf{z}_{public}^*} \in \mathcal{Y}$ and the feature vectors $\mathbf{z}^*$ obtained in the previous step. We experiment with several predictive and similarity-based algorithms. We split the public collection data $\mathbf{z}_{public}^* \in \mathcal{Z}^*$, $\mathbf{y}_{\mathbf{z}_{public}^*} \in \mathcal{Y}$ into 70\% training and 30\% validation, and predict on $\mathbf{z}_{indoor}^* \in \mathcal{Z}^*$ to obtain the fictional locations of the indoor artworks $\mathbf{\hat y}_{\mathbf{z}_{indoor}^*} \in \mathcal{Y}$.

We train a selection of canonical machine learning regression models to predict $\mathbf{y}$ using $\mathbf{z}^*$. In particular, we test two decision tree-based methods: Random Forest \cite{ho1995random}, XGBoost \cite{Chen:2016:XST:2939672.2939785}, and, a Support Vector Machine Regression (SVR) \cite{cortes1995support}\footnote{All the models used are available from \href{https://scikit-learn.org/stable/supervised\_learning.html}{sklearn}.}, using feature standardization and hyperparameter tuning, described in Appendix 1.

Moreover, we also experiment with an unsupervised GPS-inspired similarity method to compute the coordinates of the indoor artworks $\mathbf{\hat y}_{\mathbf{z}_{indoor}^*}$. We use the feature vectors $\mathbf{z}_{indoor}^*$ to find the three most similar public artworks. Practically, we construct a Ball Tree \cite{Dolatshah2015-ji} using $\mathbf{z}_{public}^*$ and we query the tree with each $\mathbf{z}_{i^*_{indoor}}$. We retrieve, for each indoor artwork, the three items of the public artworks $\mathbf{z}_{j^*_{public}}$  where $j=1,2,3$ with the smallest Euclidean distance $d$ as:

$$\mathbf{z}_{1^*_{public}} = \argmin _{i \in indoor, k \in public} d( \mathbf{z}_k^*, \mathbf{z}_i^* ) $$ 
$$\mathbf{z}_{2^*_{public}} = \argmin _{i \in indoor, k \in public} d( \mathbf{z}_k^* | \mathbf{z}_1, \mathbf{z}_i^* ) $$ 
$$\mathbf{z}_{3^*_{public}} = \argmin _{i \in indoor, k \in public} d( \mathbf{z}_k^* | (\mathbf{z}_1, \mathbf{z}_2), \mathbf{z}_i^* ) $$ 

Finally, we use the coordinates $\mathbf{y}_{\textbf{z}_{j^*_{public}}}$ of the three most similar public artworks $\mathbf{z}_{j^*_{public}}, j=1,2,3$ to triangulate the fictional coordinate of the indoor artwork as the centroid of the triangle simply as:

$$ \mathbf{\hat y}_{\textbf{z}_{i^*_{indoor}}} = \frac{\mathbf{y}_{\textbf{z}_{1^*_{public}}} + \mathbf{y}_{\textbf{z}_{2^*_{public}}} + \mathbf{y}_{\textbf{z}_{3^*_{public}}}}{3}$$

Furthermore, we test a weighted alternative of the latter method as:
$$ \mathbf{\hat y}_{\textbf{z}_{i^*_{indoor}}} = \frac{w_1 \mathbf{y}_{\textbf{z}_{1^*_{public}}} + w_2 \mathbf{y}_{\textbf{z}_{2^*_{public}}} + w_3 \mathbf{y}_{\textbf{z}_{3^*_{public}}}}{3}$$
where $w_{1,2,3}$ are defined as the complementary softmax calculation based on the Euclidean distances $d$:
$$w_j = 1 - softmax(d)[j]$$
which gives more weight and importance to the locations of artworks that are more similar to the artwork in consideration. We use the public artworks belonging to the training set to predict the final indoor artwork locations and test the performance of this method on the public artworks of the validation set.

\subsection{Fictional and Real Coordinates to Panoramas}
Once the fictional location of the indoor artworks $\mathbf{\hat y_{\mathbf{z}_{indoor}^*}}$ is calculated, we begin with the inspection of the local conditions of each data point, the 360° panorama street view. We employ the Google Street View API to gather the panorama street views $\mathbf{v}$ at each latitude and longitude tuple (both $\mathbf{y_{\mathbf{z}_{public}^*}}$ and $\mathbf{\hat y_{\mathbf{z}_{indoor}^*}}$). 

Helsinki offers a very varied landscape, spanning from coastal settings and gulfs to urban areas and parks. On the one hand, this variation provides fertile ground for the following image generation phase. On the other, the pipeline is challenged by several locations where the 360° street view is unavailable. To overcome this limitation, an iterative process queries the Google Street View API with increasing radii to ensure proximity to the predicted position, while permitting local adjustments. In the case of locations with no available street view panorama within a radius of 250 meters, such as in the middle of the sea, a 360° panorama view with an aspect ratio of 19:6 is generated using Midjourney\footnote{See \href{www.midjourney.com}{midjourney.com}.}. While for the remainder of the paper we adopt Stable Diffusion for image generation, we found that empirically Midjourney was providing better quality imaginary natural landscapes than Stable Diffusion in the absence of depth guidance.

\subsection{Panoramas and CLIP prompts to Art Panoramas}

Finally, using the panorama views $\mathbf{v}$ of each location as depth maps, and prompts $\mathbf{t}$, we generate landscape artworks that semantically depict the original art piece but use the real context as the canvas. To this end, ControlNet\footnote{We use the code from the official release on \href{https://github.com/lllyasviel/ControlNet}{Github}, v1.0.} \cite{Zhang2023-sj} plays a key role in guiding the generation with an input depth map, computed via MiDaS from $\mathbf{v}$ \cite{Ranftl2022-rr}. Through their combination - assisted with asymmetric tiling\footnote{See \url{https://github.com/tjm35/asymmetric-tiling-sd-webui/}.} - we influence the Stable Diffusion\footnote{We use the code from the \href{https://huggingface.co/runwayml/stable-diffusion-v1-5}{huggingface release} v1.5, using 30 inference steps and Euler sampling.} generation towards pertaining visual consistency between the real and the imagined landscapes. Finally, the resolution of the artwork is increased by 4x using ESRGAN \cite{Wang2019-ke}, leading to the resulting art panoramas $\mathbf{a}$ (Figure \ref{fig:6}).

\section{Evaluation}
To evaluate the various steps of the pipeline we adopt and develop both qualitative and quantitative metrics, due to the profoundly perceptual nature of the final work. Furthermore, we do not strive for a perfect prediction or generation, both with respect to the allocation of the coordinates and to the rendering of the style of the original artwork in the final panorama; oppositely, we allow for a certain degree of freedom, which, we believe, is innate to the artistic license of the work.

Particularly, we quantitatively evaluate the fictional coordinates assignment (Section 2.2 and 3.1). We set out to evaluate the results according to two criteria: spatial dispersion of the predicted coordinates, and ‘semantic’ accuracy of the new locations. The first criterion is motivated by the joint desire of our team and the HAM curatorial team to disseminate the artworks across the whole area of the city, against the natural and cultural tendency to concentrate artworks and attractions in the center or in limited areas of a city. We decided to evaluate the 2D dispersion of the predicted locations in the city using the \textit{mean Euclidean distance from the central predicted point}, quantifying a pseudo standard deviation of the points. The second criterion is related to the original idea of this AI curation: to manifest a pseudo-agnostic machine view of the city. We purposefully abstract from the specificities of the urban context and only adopt the visual and textual embeddings of the artworks to estimate a location. We then evaluate the degree of plausibility of the predicted locations. As a proxy, we use the validation set from the public artworks dataset and evaluate the methods using the \textit{mean Euclidean distance of the predicted locations of the public artworks against their real locations}. This, we believe, gives us an estimate of the predictive power of the two methods when given as input only the visual-textual embeddings and none of the urban confounding factors.

To evaluate the quality of the final panoramas we use both qualitative and quantitative methods (Section 3.2). Qualitative assessment is in essence highly subjective, especially in the context of artistic creation, which makes it challenging to evaluate. When starting this project, the team had in mind specific requirements regarding the output produced by the computational process, which are related to the level of abstraction of the created image, the degree of visual similarity with the original artworks, and the sense of connection with the urban environment where the artwork is projected. Furthermore, the project was developed with periodic feedback from people directly involved with the HAM collection, which consists of a team of expert curators. We assess the opinion of this expert public, as well as integrate a quantitative evaluation to support the qualitative aspects taken into account. 
We perform the qualitative assessment in the form of a questionnaire, which presents multiple-choice, ordinal, and interval scale questions. The survey is conducted with a group of expert curators based in the city of Helsinki and are evaluating the production of AI curation on the following criteria:
\begin{enumerate}
  \item the level at which the essence of the original panorama is retained in the final panorama,
  \item the level at which the essence of the original artwork  is preserved in the generated panorama,
  \item the strength of the relationship between the original artwork and the assigned urban context.
\end{enumerate}
 
The questionnaire is detailed in Appendix 2.

The quantitative assessment of the produced artwork addresses, partially, the structural similarity between the original panoramas and the generated panoramas.


To evaluate the extent to which the the urban landscapes are preserved in the computationally generated art panoramas we use two visual similarity metrics. To avoid unfair raw pixel comparisons, we first capture perceptual similarity using Structural Similarity (SSIM), which encodes further subtle properties of images such as structural information. Since SSIM still relies solely on pixel intensities, we reduce the influence of pixels that unlikely contain structural information by computing the SSIM metric of the Canny Edges images instead of the original images. Secondly, we use Histogram of Oriented Gradients (HOG) to build feature vectors that encode shape, edge patterns and texture details from the images. We compute the distribution of orientations in each 16x16 patch with 8 possible orientations for each image and use these to compare the line dynamics between the original panorama and the generated one.


\section{Results}
In this section, we present our quantitative and qualitative results, related to the computation of fictional coordinates and the quality of the generated panoramas. 

\subsection{Fictional coordinates}
\begin{table}[h!]
\centering
\begin{tabular}{l|r|r|r|r|r|}
\cline{2-6}
                                          & \textbf{SVR} & \textbf{Random Forest} & \textbf{Extreme Gradient Boosting} & \textbf{Similar} & \textbf{Similar Weighted} \\ \hline
\multicolumn{1}{|r|}{Mean Absolute Error} & \textbf{0.028} & 0.030                  & 0.030                              & 0.036            & 0.036                     \\ \hline
\multicolumn{1}{|r|}{Mean Squared Error}  & 0.002          & \textbf{0.001}         & 0.002                              & 0.002            & 0.003                     \\ \hline
\multicolumn{1}{|l|}{Dispersion}          & 0.094          & 0.173                  & 0.288                              & 0.444            & \textbf{0.446}            \\ \hline
\end{tabular}
\caption{A table presenting the results of the evaluation of the three machine learning-based method (SVR, Random Forest, XGBoost) and the two variations of the Similarity-based methods (Similar and Similar Weighted) at the fictional coordinate prediction. The Mean Absolute and Mean Squared Error refer to the second method in Section 3, assessing the 'semantic' accuracy of the model predictions; the Dispersion refers to the first method in Section 3, the geographic dispersion of the predictions in the city, quantified as the mean Euclidean distance from the central predicted point.}
\label{tab:results}
\end{table}

\begin{figure}[h!]
  \includegraphics[width=0.45\textwidth]{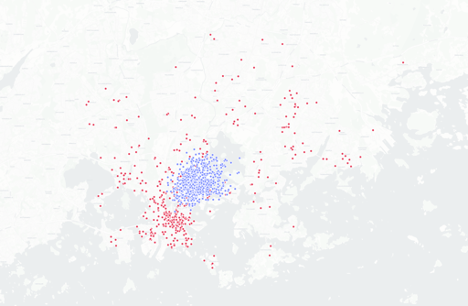}
  \includegraphics[width=0.45\textwidth]{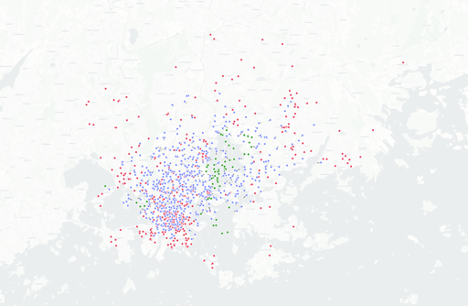}
  \caption{Top: Map of Helsinki showing the public artworks $\mathbf{y}_{\mathbf{z}_{public}^*}$ in red, and the predicted locations $\mathbf{\hat y}_{\mathbf{z}_{indoor}^*}$ in blue. Clustered blue points illustrate how the Random Forest model does not capture variation in the data. Bottom:
  Updated map with the GPS-inspired similarity method. Green points indicate sea locations that did not have a 360° panoramic image.}
  \Description{}
  \label{fig:34}
\end{figure}

We compare and evaluate the performance of the assignment of the fictional coordinates as explained in Section 3. The results are presented in Table \ref{tab:results}. After the hyperparameter tuning, the best-performing models with respect to the mean squared and absolute error in the predictions with respect to the original coordinates are the machine learning-based methods, with a marginal preference for the SVR in terms of absolute error and Random Forest for squared error. On the other side, the spatial dispersion of the machine learning-based models is rather poor, with a great improvement in the dispersion happening in the similarity-based methods. The geographic dispersion is also visible in Figure \ref{fig:34}, where it becomes clear that the predictions of the Random Forest are not able to capture the variance of the original data. 

The similarity method overcomes such low prediction variation, with only a slight increase in the error rates (which is almost none for the non-weighted similarity method). This approach results in a roughly even and well-distributed set of locations, which are easy to navigate and distinctive from each other (Figure \ref{fig:34}). 

As there are no ground truth locations for the indoor artworks, we are only interested in a meaningful way of representing the locations through the eyes of the machine. We aim to recreate the space of the city, an alternative Helsinki through the lens of machine perception. The unsupervised method is justified exactly by this premise, that the space should follow a semantic continuity rather than a strictly geographical one. In addition, the triangulation is inspired by GPS localization, reconnecting symbolically to the bases of geolocation. 
      
Therefore, we select GPS-inspired similarity as the most suitable method to generate fictional locations. We prefer the non-weighted similarity over the weighted option despite a slight decrease in dispersion because of the marginal improvement in the squared error. With this method, we obtain a dataset of 1’744 predicted locations of which we are able to retrieve 1’681 equirectangular street view images within a range of 250 meters. The remaining 3.61\% are assigned the set of generated HDR (High Dynamic Range) images of forest (14\%) and sea (86\%) landscapes. 

\begin{figure}[h!]
  \includegraphics[width=0.5\textwidth]{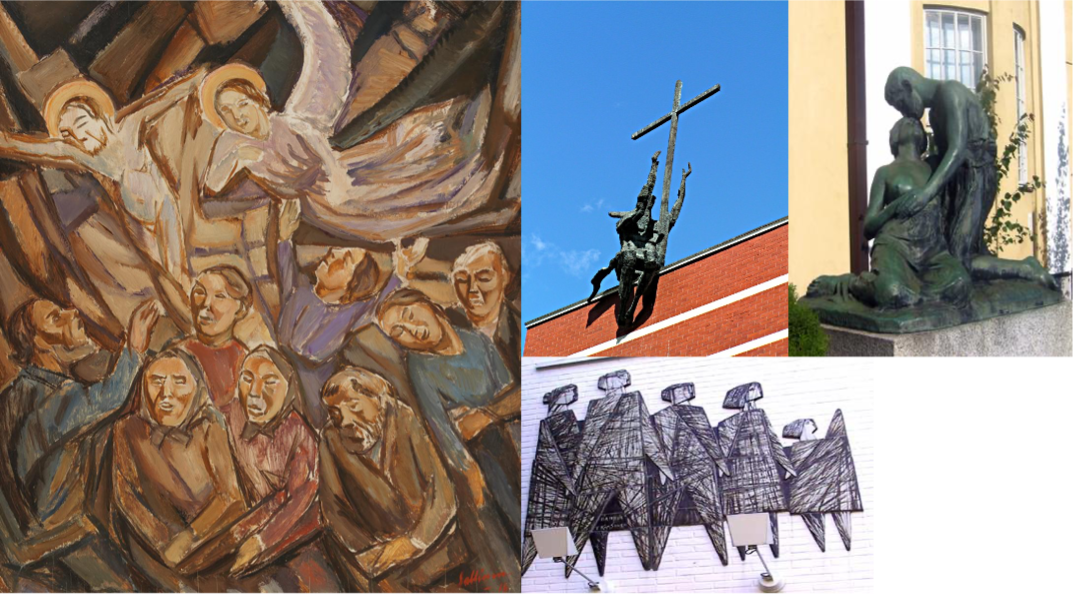}
  \caption{Example of the top three most similar public artworks (right) to an indoor painting sample (left). Artworks (from left to right top to bottom): Sallinen, T. "Hihhulit"; Juva, K. "Arkkienkeli Mikael"; Sörensen-Ringi, H. "Jäähyväiset"; Kaasinen, T. "Ihmisiä". Courtesy of the HAM collection.}
  \Description{}
  \label{fig:5}
\end{figure}

 We inspect the three most similar images from a randomly selected set of indoor artworks from which we draw some general impressions on the nature of such similarity. The retrieved images seem to have both conceptual and visual similarities to the original image (Figure \ref{fig:5}), where the concept of religion is used in the retrieval of the first image, even in the absence of any physical connection. In some cases, similarities are found across modalities, with instances of connections between paintings and statues. For example, formalistic properties are shared between query and retrieved images (e.g., a drawing of a square triggers cubic-shaped sculptures in retrieval). On the other hand, some public artworks are retrieved repeatedly, sometimes without any clear connection, indicating that the search space is not equally likely for all images. The limited conceptual connections are unsurprising as CLIP prompts function as textual descriptions of the visual, and not as an art historical explanation of the artwork. Moreover, we note that CLIP-Interrogator only adopts a limited vocabulary in the extraction of the textual description, which is not suited for art historical descriptions. 

\subsection{Art Panoramas}
Results for the evaluation of the art panoramas in relation to the original panorama are presented in table \ref{tab:results_vissim}. We present a statistical summary of the computed metrics between every original-art panorama pair. SSIM score ranges from 0 to 1, with 1 indicating a perfect similarity between the images, and 0 indicating no perfect mismatch. The HOG metric is computed using the cosine similarity of the extracted HOG feature vectors, with 1 indicating perfect similarity and -1 complete dissimilarity. 

The expected value of the SSIM metric suggests that while some aspects of the structure are preserved, there are noticeable variations introduced by the artistic transformation. Metrics show that using edge information reduces the influence of pixels that do not contain information about urban compositions displayed on the images. This is shown by the improvement in the overall SSIM score when edge image information is used as input (SSIM Edges). Images that are characterised by lower frequency patterns tend to have higher SSIM scores, while high frequency patterns are responsible for the lowest socres. For example, an art panorama showing \textit{puntillism} style is less likely to share structural patterns with respect to the original street view. In contrast, low frequency patterns (e.g., a clear sky, an empty road) influenced by a variety of artistic styles that maintain such low frequency patterns, are translated into higher similarity scores (see Figure \ref{fig:6}). The higher value of the proposed HOG (8) suggests that this metric is less sensitive to undesired visual artefacts present in the original image, such as clouds, shadows, and other visual elements that do not encode structural information of the input image. The fact that HOG (8) metric is less sensitive to these artefacts is reflected by a lower standard deviation, suggesting that values are tightly clustered around the mean, and a significant higher minimum score. Metrics are influenced by both, variability of the original street view, and variability of the artistic style. Qualitative inspection of obtained results seem to suggest that the simpler the images, the more chances to maintain structural similarity between the original and artistic panorama. In short: Image complexity penalises the proposed metrics and serves us as an indicator for structural similarity comparison.

\begin{table}[]
\centering
\begin{tabular}{l|l|l|l|l|}
\cline{2-5}
                                 & \textbf{Mean} & \textbf{Std} & \textbf{Min} & \textbf{Max} \\ \hline
\multicolumn{1}{|l|}{\textbf{SSIM}}       & 0.369    & 0.121   & 0.027   & 0.731  \\ \hline
\multicolumn{1}{|l|}{\textbf{SSIM Edges}} & 0.421    & \textbf{0.156}   & \textbf{0.001 }  & \textbf{0.854}   \\ \hline
\multicolumn{1}{|l|}{\textbf{HOG (8)}}    & 	\textbf{0.651}    & 0.052   & 0.366   & 0.766   \\ \hline
\end{tabular}
\caption{Statistical summary of relevant metrics used to evaluate performance between every original-art panorama pair in the dataset.}
\label{tab:results_vissim}
\end{table}

\begin{figure}[h!]
  \includegraphics[width=1\textwidth]{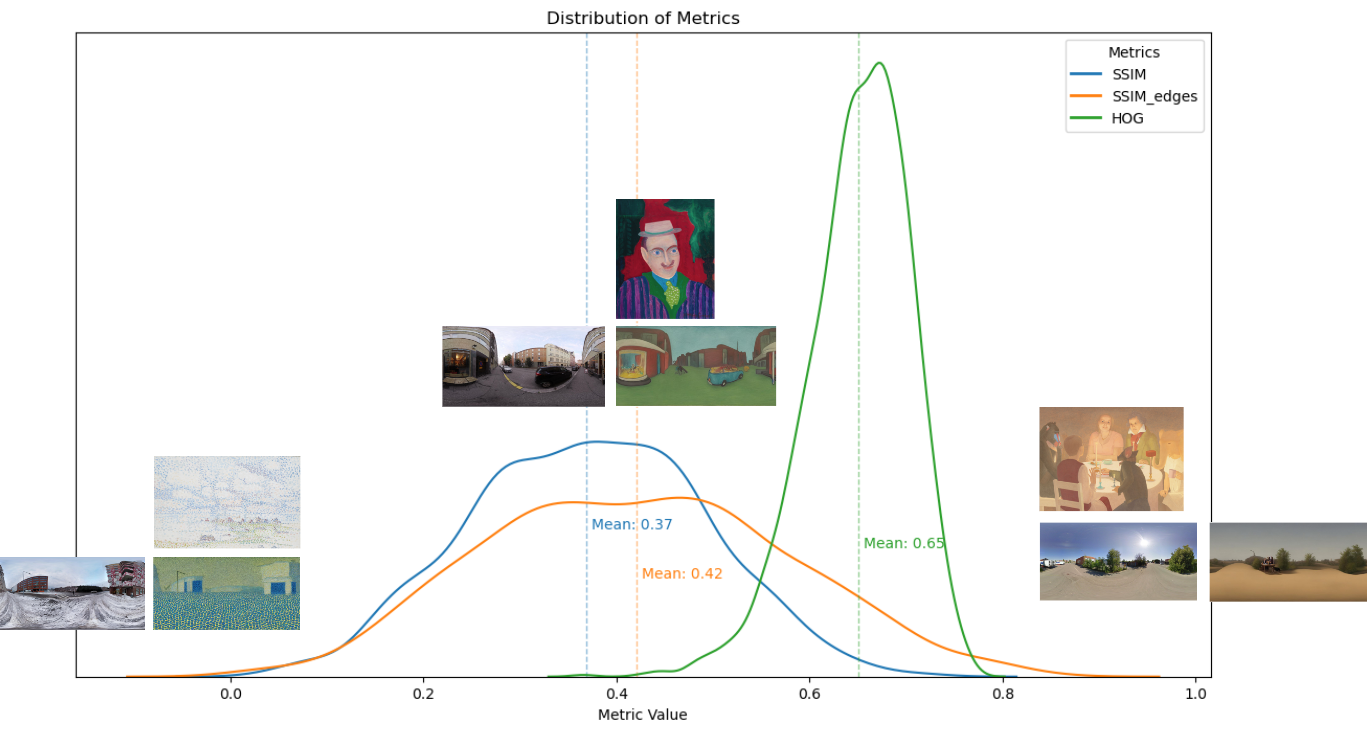}
  \caption{Distribution of the proposed metrics. 3 image-pair examples are showed from low, medium, and high scores. Each of them showing the original painting above, the original panorama to the left and the art panorama to the right.}
  \Description{}
  \label{fig:6}
\end{figure}

Furthermore, observing the panoramas ourselves, we see that in most cases geometry is kept and the perceived resulting space still captures the 3-dimensional quality of the original urban setting (Figure \ref{fig:5}), the results show that the built physical features can become highly transformed depending on the graphical and pictorial style of the source artwork. For instance, very atmospheric pictorial styles, where contours are very blurred and diffuse, tend to generate resulting 360° art panoramas with less recognizable geometries of specific physical features of the original urban elements, while keeping the same perception of depth and overall composition. 

We further observe that most of the panoramas broadly reflect high coherence with respect to the semantics but significant shifts in color palette and brushstroke. Concretely, we highlight how the textual information used to generate art panoramas is often insufficient for an appropriate match of the color palette and style properties of artworks. The perceived space changes with the artistic style of the original artwork, rendering a new dreamed Helsinki as seen through the collection by the proposed pipeline.

The immersion of the resulting 360° art panoramas is generally satisfactory as tested through a generic online 360° panorama viewer VR\footnote{See \href{https://renderstuff.com/tools/360-panorama-web-viewer/}{renderstuff.com}.}. It is to be noted that those panoramic views are experienced through a fixed standpoint and do not each result in a synthetic 3D navigable space. Nevertheless, using 360° panoramas allows us to capture a complete spherical view of the image surroundings, hence leading users towards a credible immersive experience. 

\begin{figure}[h!]
  \includegraphics[width=0.5\textwidth]{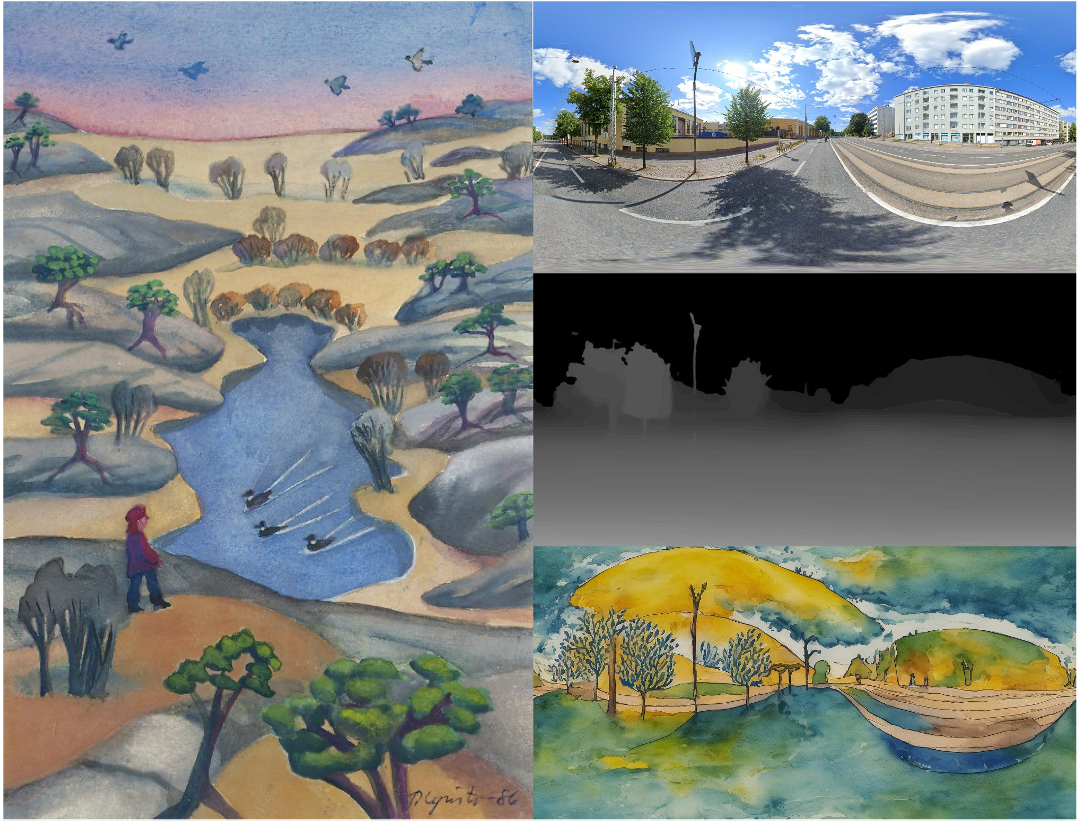}
  \caption{Images involved in the immersive panorama generation of Figure \ref{fig:1} (left). Panorama of predicted location (top right), Depth map (middle right), art panorama using depth map and CLIP prompt (bottom right). Artwork: Lepistö, P. "Kolme vesilintua metsälammella". Courtesy of the HAM collection.}
  \Description{}
  \label{fig:6}
\end{figure}

\section{Discussion}
Previous sections have presented our curatorial work for the 2023 Helsinki Biennial. The process of creating a new spatial projection of artworks from the HAM collection entails ethical and scholarly issues. These are part of a larger narrative that continuously unfolds and is focused on the ways creative AI applications (en)force global cultures. 

\subsection{Towards a curatorial machine}
The stack of models presented works as a curatorial agent, not simply following another sophisticated search engine paradigm, but as a mediating actor. In modern society, we are already witnessing machine curatorial processes that guide cultural aesthetic preferences \cite{Manovich2017-ln} (e.g., recommender systems), but this example showcases and highlights new emergent practices that point to a shift in the modalities involved in cultural curation and artistic production. Particularly, agency comes into question in this work, a modular stacking of several algorithms from a variety of tasks and applications. The complexity of such products makes it unsurprising that both scholars and the broader public are ready to project authorship to these models as we have recently seen in debates about ChatGPT and GPT3-4 \cite{Lund2023-mn, Sample2023-gl, Zhuo2023-np}. In our case, the capacities of a curatorial agent which relate to the possibility of connecting images and texts to a common embedding space, are inherited from CLIP. Furthermore, the capacity to establish translations between the visual, textual, and spatial, situates the imagined collection in a new geography that is neither unreal nor illusory as it is shared and can be experienced, much like curation.  Curation is no longer reserved for humans, just as this curatorial practice transgresses the domain-specific knowledge of the art world. As machine curation develops and normalizes, understanding the inner functioning of machine learning models becomes an increasingly crucial literacy, relevant to a general set of skills in contemporary artistic curation.

\subsection{Avenues for public engagement}
As this curation is not a physical exhibition, it can inhabit endless physical spaces, each creating a diverse public engagement. The first possibility of physical interaction is inside the walls of the HAM. This can take the form, among others, of a 2D projection screen, a 3D immersive curved screen, or a Virtual Reality headset. Due to the nature of the panoramas, we believe the best interaction would be achieved through a cylindrical screen or a dome, which would immerse the public in a 360 degrees setting of the surrounded mutated city, and, unlike the VR headset, this projection is suited for the fixed central standpoint of the produced panoramas. The geographical nature of the project opens a second avenue for physical interaction outside the museum space: the street. Here, we believe an interesting interaction can take place as a mobile AR experience. The public would engage with the machinic Helsinki at the location of the real Helsinki, superimposing the two worlds simultaneously. 

These avenues lead us to ask what the effects of this newly imagined urban landscape on the “real” Helsinki are, and what interactions it will unleash once deployed. The impact of digital versions of a given environment, especially urban, is being discussed within the digital twin and smart city scene \cite{Lv2022-ze}, but the effects of such an imagined digital version on the behaviors, attitudes, urban development, and other aspects of the social and built urban environment are yet to be assessed. 

In addition, the potential effects of such an experience in fostering enhanced levels of spatial agency are also to be considered and assessed. Aesthetic experiences can be conducive to emotional reactions, which in turn alter not only the way we perceive the space around us but also can modulate our perception of affordances and therefore, our spatial capacities \cite{negueruela2017city}. Our project seeks to revisit the urban spaces of Helsinki through the imagined panoramas, inviting the public to engage differently with their urban imaginary. An AR implementation that locally reacts to the location of the visitors, either through geolocation estimation via GPS triangulation or through on-site QR code scanning can add a crucial component, resulting in a more convincing situated experience and therefore, a more impactful emotional and affective engagement.

\subsection{Ethical considerations}
In our globalized world, an art biennial is an event designed to showcase and locate a city on the map of current creative and influential cities worldwide. It is, therefore, about status and attracting attention, explicitly foregrounding the added value of the city’s assets, innovative profile, and capacity to become a central player \cite{Kompatsiaris2017-gu}. In the case of this project, we were commissioned to work with and feature the collections of the HAM collection, which is primarily composed of artworks from local artists, potentially for a global audience. We tackled such a conundrum with the clear objective of avoiding manipulating the original artworks in order to respect their artistic integrity and their rich and complex relationship with the many layers of memory, identity, and local cultural heritage. 

The ethical stakes of such an operation need to be carefully delineated. On the one hand, there is a need to respect, as outlined above, the artwork regarding its cultural context, and, on the other hand, we need to consider the approach from a global public, necessarily agnostic of those specificities. This second approach requires some freedom to appropriate, recombine, and re-imagine cultural production in the process of international artistic influence and cross-fertilization. 
    
In this conundrum, the machine (meaning the actant resulting from the stack of diverse models) becomes an aid. It allows a certain lecture of these images. However, the fact that we are using a CLIP-guided model, and getting the textual and visual embeddings from CLIP means that we are, in practice, reading a collection (the one of HAM) from another collection (the CLIP training dataset). What that exactly means remains a complex and multifaceted question, but in our case, it becomes a crucial aspect that problematizes the desired cultural agnostic approach as made through the machine. It confirms that cultural framing is already embedded in our computational models and begs the question of how to conceive an AI curation that embraces diverse cultural frames and a non-colonial approach.

\section{Conclusion}
In this paper, we presented a novel AI curation of a collection of works of art carried out for the 2023 Helsinki Biennial. Because of our will to propose a culturally agnostic view on artistic material, we decided to represent the collection through the lens of deep learning models, proposing a machinic approach to art curation, and anchoring it in the city of Helsinki. Using the Helsinki Art Museum (HAM) collection, we considered the use of the CLIP-Interrogator to create textual descriptions of the works of art and assign them fictional coordinates around the city of Helsinki through a similarity-based algorithm. A new synthetic 360° panoramic view of the predicted location was then generated with the original depth map of the location and the CLIP prompt, thus proposing a new visual style or flavor of the city of Helsinki. We introduced a discussion on the current AI curation practices and the innovations presented by our work. Moreover, we elaborated a comprehensive evaluation exploiting descriptive, quantitative, and qualitative methods. 

Part of the generated material is made accessible with the implementation of a web application in collaboration with the designer Yehwan Song\footnote{See \url{https://yhsong.com/}.}. The goal of the platform is to propose to the user the possibility to navigate these fictional projections, to move from one space to the next, thus discovering the HAM collection through the gaze of the machine and opening the question of the structure of the navigation in this synthetic geography. As a complement to this free navigation, a future line of research will be to use other machine learning models to automatically generate narratives and shape new threads of exploration of that space.



\begin{acks}
Digital Visual Studies is a project funded by the Max Planck Society. Furthermore, we would like to thank the team behind the organization of the 2023 Helsinki Biennial as well as those responsible for the Helsinki Art Museum collections for their collaboration and assistance. We would also like to thank the artists featured in the HAM collections whose artworks constitute the source materials for the project.

\end{acks}

\bibliographystyle{ACM-Reference-Format}
\bibliography{main}

\appendix
\section{Hyperparameter Tuning}
\textbf{Searching for the optimal parameters}. To ensure that our results are not dependent on the hyperparameter settings of the regression models and to obtain optimal results, we carry out a randomized-search-based hyperparameter tuning  with cross-validation on the three machine learning regression models we use (Random Forest, XGBoost, and Support Vector Machine Regression). We test 10 settings for 3 data splits for each model, totalling 90 configurations. The best configuration for each model is chosen based on the \textit{mean absolute error (MAE)}.

\subsection{Random Forest}
For the Random Forest regression we tune the following parameters:  \\
\texttt{n\_estimators: 200, 400, 600, 800, 1000, 1500.} \\
\texttt{max\_depth: 5, 10, 15, 40, 50, 55}  \\
\texttt{max\_features: 1.0, sqrt, log2}  \\
\texttt{criterion: squared\_error, absolute\_error}  \\
\texttt{min\_samples\_split: 2, 5, 6, 7, 9, 10}  \\
\texttt{min\_impurity\_decrease: 0.01, 0.05, 0.1}  \\
\texttt{bootstrap: True, False}.  \\

\begin{table}[ht]
\centering
\begin{tabular}{@{}llllllll@{}}
\toprule
\textbf{n\_estimators} & \textbf{min\_samples\_s} & \textbf{min\_imp\_dec} & \textbf{max\_features} & \textbf{max\_depth} & \textbf{criterion} & \textbf{bootstrap} & \textbf{MAE} \\ \midrule
800                    & 6                            & 0.05                             & sqrt                   & 55                  & squared\_error     & False              & 0.033                        \\
600                    & 10                           & 0.01                             & log2                   & 10                  & squared\_error     & True               & 0.033                        \\
600                    & 9                            & 0.01                             & 1.0                    & 5                   & absolute\_error    & True               & \textbf{0.0311}              \\
200                    & 7                            & 0.1                              & sqrt                   & 15                  & squared\_error     & False              & 0.033                        \\
1000                   & 10                           & 0.05                             & 1.0                    & 10                  & squared\_error     & False              & 0.033                        \\
400                    & 2                            & 0.1                              & log2                   & 10                  & absolute\_error    & True               & 0.0311                       \\
600                    & 2                            & 0.01                             & log2                   & 10                  & absolute\_error    & False              & 0.0311                       \\
1000                   & 10                           & 0.01                             & 1.0                    & 50                  & squared\_error     & True               & 0.033                        \\
1500                   & 2                            & 0.01                             & log2                   & 40                  & squared\_error     & False              & 0.033                        \\
1500                   & 5                            & 0.01                             & sqrt                   & 10                  & squared\_error     & False              & 0.033                        \\ \bottomrule
\end{tabular}
\caption{Hyperparameter results for Random Forest configurations.}
\label{tab:rf}
\end{table}
        
For more information on the meaning and options of the hyperparameters please refer \href{https://scikit-learn.org/stable/modules/generated/sklearn.ensemble.RandomForestRegressor.html}{here}. The results of the hyperparameter tuning are shown in Table \ref{tab:rf}. We observe very little variation in the results, which a slight advantage of the methods using absolute\_error as criterion. We therefore select the parameters: 600, 9, 0.01, 1.0, 5, absolute\_error, True.

\subsection{XGBoost}
For the XGBoost regression we tune the following parameters:  \\
\texttt{n\_estimators: 800, 1200, 1400, 1600, 2000} \\
\texttt{max\_depth: 2, 4, 10, 14, 16, 20}  \\
\texttt{min\_child\_weight: 2, 3, 4, 5, 9, 10}  \\
\texttt{objective: reg:squarederror, reg:squaredlogerror}  \\
\texttt{tree\_method: exact, approx, hist}  \\
\texttt{eta: 0.1, 0.2, 0.3, 0.4, 0.5}.  \\

For more information on the meaning and options of the hyperparameters please refer \href{https://scikit-learn.org/stable/modules/generated/sklearn.ensemble.GradientBoostingRegressor.html#sklearn.ensemble.GradientBoostingRegressor}{here}. The results of the hyperparameter tuning are shown in Table \ref{tab:xgb}. We observe that most models do not perform well, with the exception of those using squarederror as objective. We therefore select as optimal parameters: approx, reg:squarederror, 800, 10, 4, 03.

\begin{table}[ht]
\centering
\begin{tabular}{@{}lllllll@{}}
\toprule
\textbf{tree\_method} & \textbf{objective}  & \textbf{n\_estimators} & \textbf{min\_child\_weight} & \textbf{max\_depth} & \textbf{eta} & \textbf{MAE} \\ \midrule
hist                  & reg:squaredlogerror & 1200                   & 10                          & 4                   & 0.3          & 34.84                        \\
approx                & reg:squaredlogerror & 1400                   & 9                           & 14                  & 0.3          & 34.05                        \\
approx                & reg:squarederror    & 800                    & 10                          & 4                   & 0.3          & \textbf{0.03}                         \\
hist                  & reg:squaredlogerror & 1600                   & 2                           & 20                  & 0.5          & 26.27                        \\
hist                  & reg:squaredlogerror & 1400                   & 3                           & 2                   & 0.2          & 29.86                        \\
approx                & reg:squarederror    & 2000                   & 9                           & 16                  & 0.4          & 0.03                         \\
exact                 & reg:squaredlogerror & 1800                   & 9                           & 10                  & 0.1          & 34.79                        \\
approx                & reg:squaredlogerror & 800                    & 4                           & 16                  & 0.3          & 31.05                        \\
hist                  & reg:squaredlogerror & 2000                   & 5                           & 16                  & 0.2          & 32.31                        \\ \bottomrule
\end{tabular}
\caption{Hyperparameter results for XGBoost configurations.}
\label{tab:xgb}
\end{table}

\subsection{Support Vector Regressor}
For the Support Vector regression we tune the following parameters:  \\
\texttt{degree: 0, 1, 2, 3, 4} \\
\texttt{kernel: linear, poly, rbf}  \\
\texttt{C: 0.01, 0.05, 0.41, 0.81, 1.2, 1.6, 2.0}  \\
\texttt{epsilon: 0.01, 0.11, 0.3, 0.4, 0.5}.  \\

For more information on the meaning and options of the hyperparameters please refer \href{https://scikit-learn.org/stable/modules/generated/sklearn.svm.SVR.html#sklearn.svm.SVR}{here}. The results of the hyperparameter tuning are shown in Table \ref{tab:svr}. We observe that the best results are obtained with poly kernel, 0.01 epsilon, 1 degree, and C 0.05.

\begin{table}[ht]
\centering
\begin{tabular}{@{}lllll@{}}
\toprule
\textbf{kernel} & \textbf{epsilon} & \textbf{degree} & \textbf{C} & \textbf{MAE} \\ \midrule
linear          & 0.11             & 4               & 0.01       & 0.06                         \\
poly          & 0.01              & 1               & 0.05        & \textbf{0.028}                         \\
poly            & 0.3              & 3               & 0.01       & 0.08                         \\
linear          & 0.11             & 4               & 2.0        & 0.06                         \\
linear          & 0.4              & 4               & 0.41       & 0.08                         \\
linear          & 0.01             & 1               & 1.2        & 0.06                         \\
linear          & 0.3              & 2               & 1.6        & 0.08                         \\
linear          & 0.01             & 2               & 0.01       & 0.06                         \\
rbf             & 0.01             & 0               & 0.81       & 0.05                         \\
poly            & 0.5              & 1               & 1.6        & 0.08                         \\ \bottomrule
\end{tabular}
\caption{Hyperparameter results for SVR configurations.}
\label{tab:svr}
\end{table}

\section{Qualitative Feedback Questionnaire}
\begin{figure}[h!]
  \includegraphics[width=0.45\textwidth]{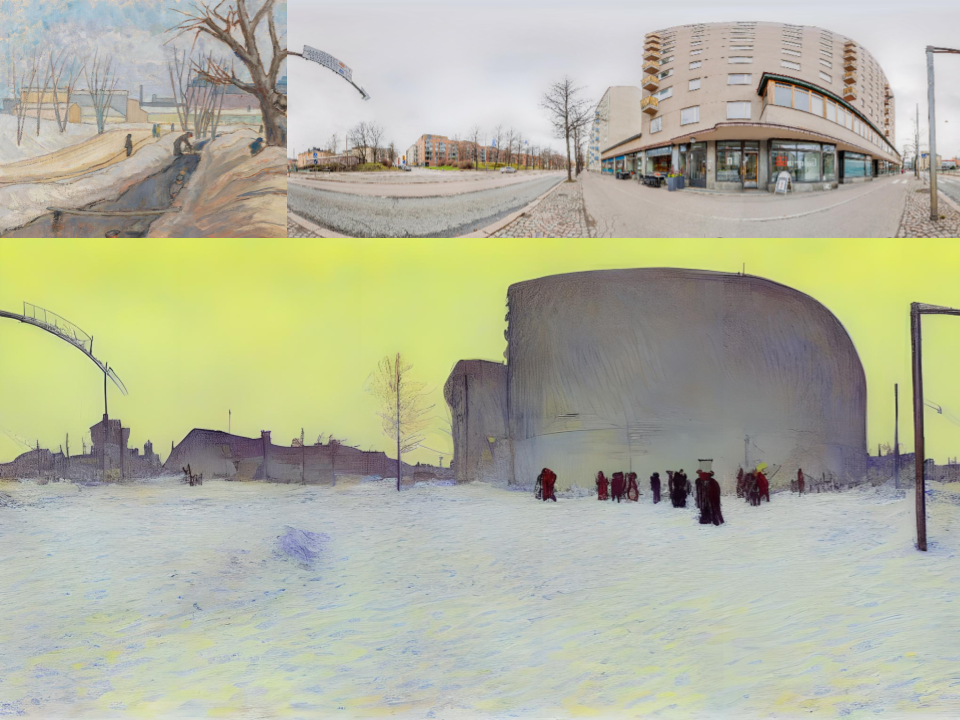}
  \includegraphics[width=0.45\textwidth]{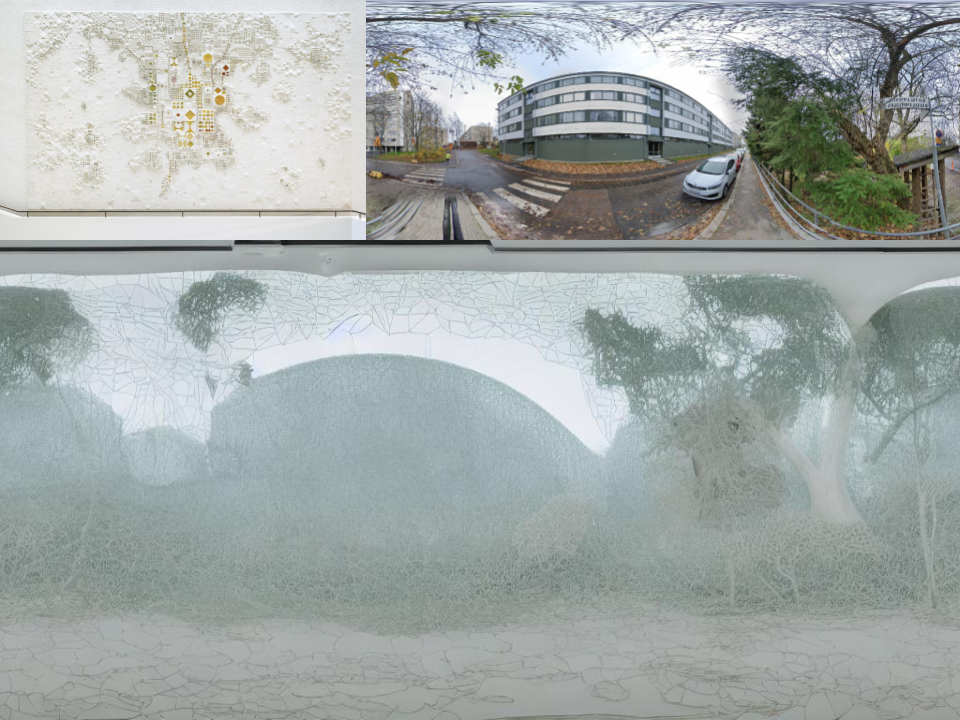}
  \includegraphics[width=0.45\textwidth]{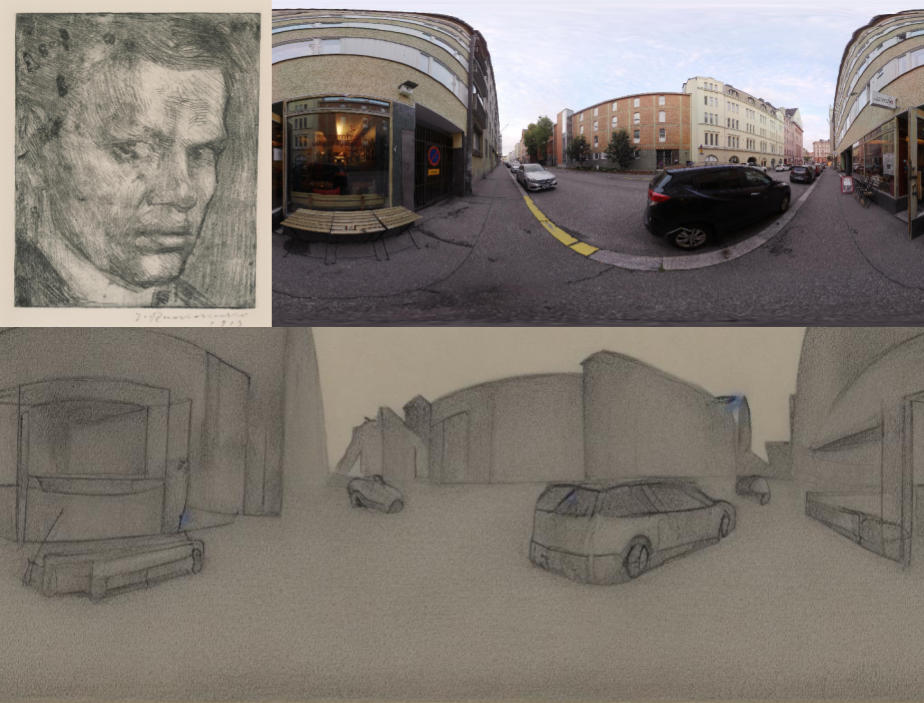}
  \includegraphics[width=0.45\textwidth]{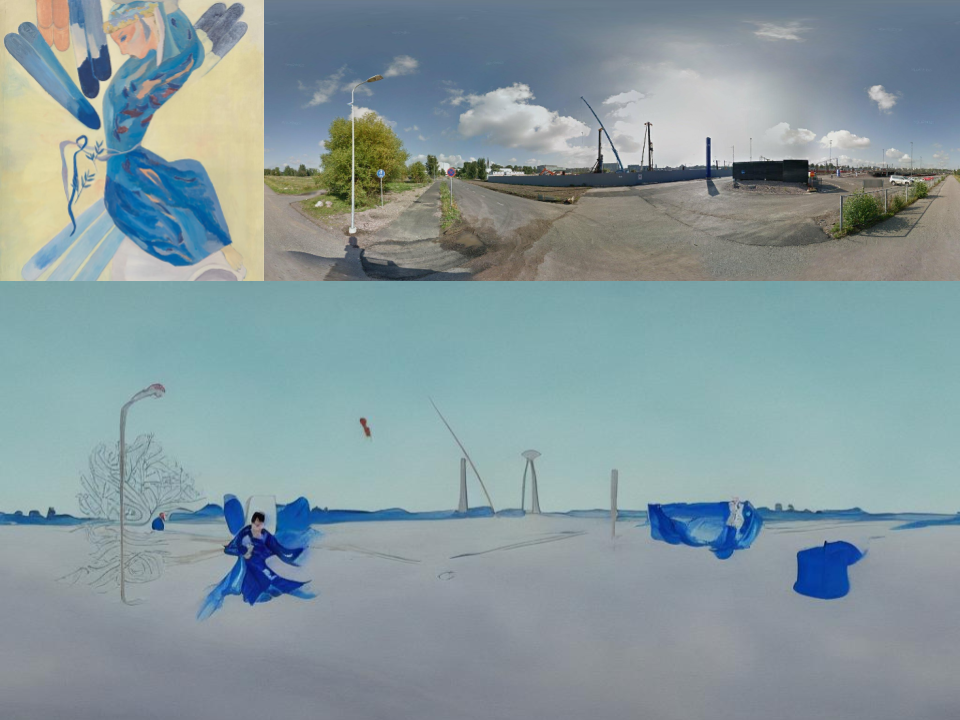}
  \includegraphics[width=0.45\textwidth]{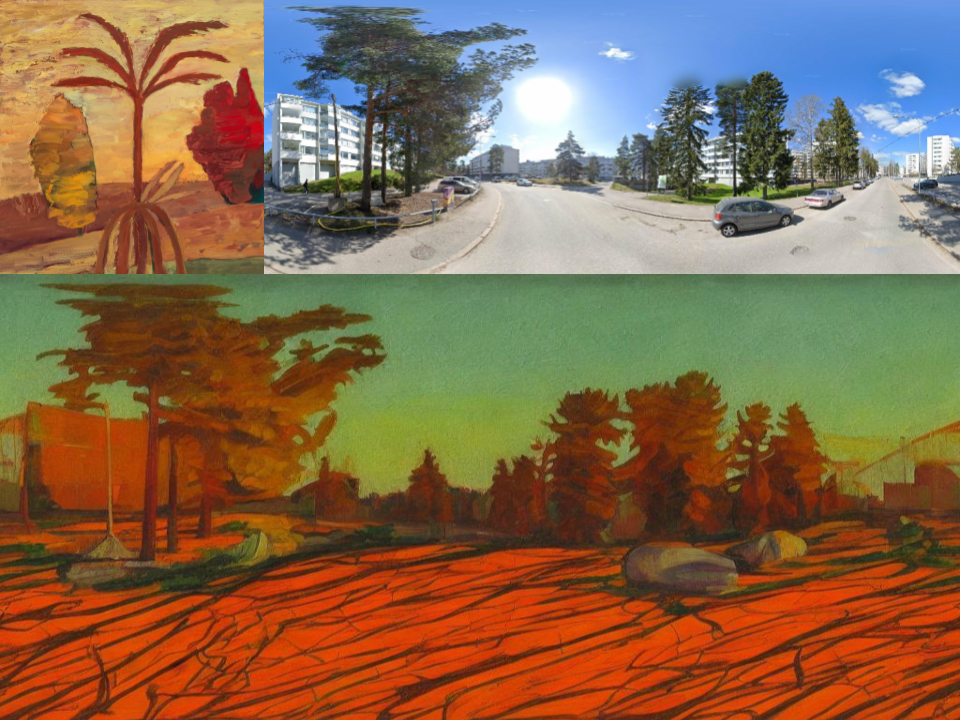}
  \includegraphics[width=0.45\textwidth]{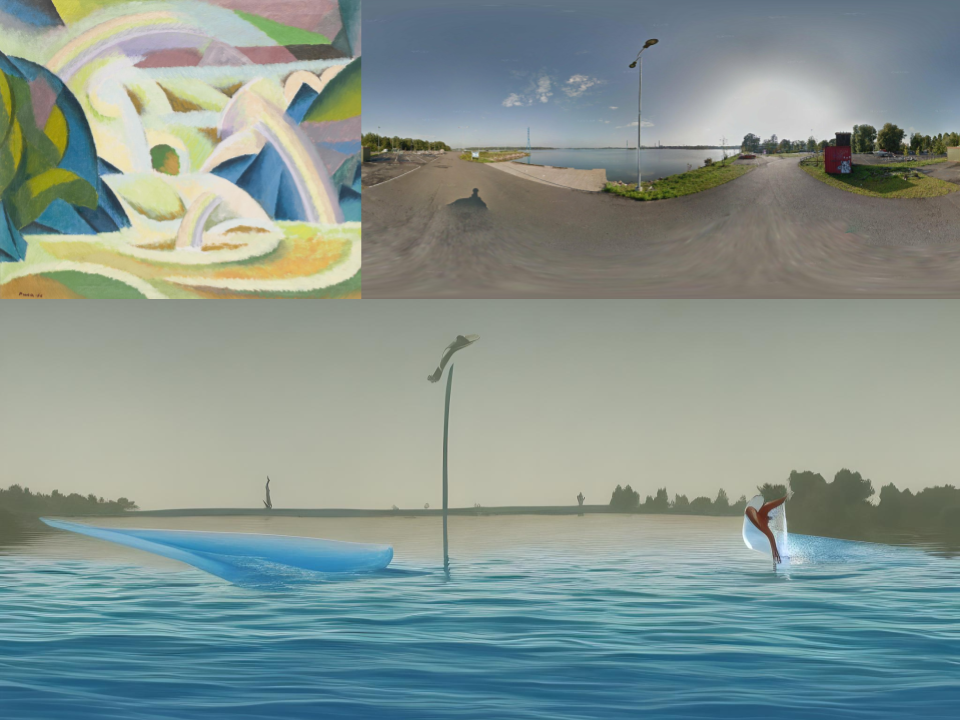}
  \caption{Sample images presented in Questionnaire. In each image, the original artwork is on the top left, the original panorama of the predicted location is on the top right and the bottom contains the final generated panorama.}
  \Description{}
  \label{fig:7}
\end{figure}

\textbf{Overview}. We selected 6 random sets of transformations, from the original image, the panorama at the predicted location and the final transformed panorama (in Figure \ref{fig:7}), and we evaluate, for each image the perceived level of coherence (between the original image and the final panorama, between the final panorama and original panorama, and between the original image and panorama, the last one being rather speculative). We use a 5 point scale ranging from 'Nothing from the original image / panorama can be found' to 'The generated image and the original seem to belong to each other'.

\textbf{Questions}

For each image:
\begin{enumerate}
  \item \textbf{Q1} How much of the essence of the original 360° panorama street view (on the top right) is preserved in the generated image (on the bottom)?
  \item \textbf{Q2} Overall, how much of the essence of the original artwork (on the top left) is preserved in the generated panorama (on the bottom)?
  \item \textbf{Q3} Do you feel that the original artwork (top left) has a relation to the urban environment (top right)?
\end{enumerate}

Other questions:
\begin{enumerate}
  \item \textbf{Q4} What criteria do you use to estimate how much of the original artwork is preserved? (colors, shapes, brush strokes, general structure, mood, etc...)
  \item \textbf{Q5} If you understand why an artwork is linked to a urban environment, what are the elements that make you think so? (visual links, knowledge on the original artworks or the city of Helsinki, types of urban landscape)
\end{enumerate}
\end{document}